\documentclass{article}

\usepackage[english]{babel}

\usepackage[letterpaper,top=2cm,bottom=2cm,left=3cm,right=3cm,marginparwidth=1.75cm]{geometry}

\usepackage{amsmath}
\usepackage{graphicx}
\usepackage{authblk}
\usepackage[colorlinks=false]{hyperref}
\usepackage[labelformat=empty, position=top]{subcaption}
\usepackage[sorting=none]{biblatex}

\title{Evidence from counterfactual tasks supports emergent analogical reasoning in large language models}

\author[1,*]{Taylor Webb}
\author[1]{Keith J. Holyoak}
\author[1,2]{Hongjing Lu}
\affil[1]{Department of Psychology}
\affil[2]{Department of Statistics\protect\\University of California, Los Angeles, CA, USA}
\affil[*]{Correspondence to: taylor.w.webb@gmail.com}

\date{}

\begin{document}

%TC:ignore

\maketitle

\begin{abstract}

We recently reported evidence that large language models are capable of solving a wide range of text-based analogy problems in a zero-shot manner, indicating the presence of an emergent capacity for analogical reasoning~\cite{webb2023emergent}. Two recent commentaries~\cite{hodel2023response,lewis2024using} have challenged these results, citing evidence from so-called `counterfactual' tasks in which the standard sequence of the alphabet is arbitrarily permuted so as to decrease similarity with materials that may have been present in the language model's training data. Here, we reply to these critiques, clarifying some misunderstandings about the test materials used in our original work, and presenting evidence that language models are also capable of generalizing to these new counterfactual task variants.

\end{abstract}

%TC:endignore

\section{Original test materials}

Hodel \& West (HW)~\cite{hodel2023response} argue that the test problems employed in our original paper~\cite{webb2023emergent} may be similar to materials available on the internet, and that such overlap may explain the ability of GPT-3 and GPT-4 to solve these problems. This argument, however, ignores the fact that many of the tests employed in the original study were designed from scratch, specifically for the purpose of testing language models on unusual problems dissimilar to materials commonly found on the internet. Most importantly, our tests included a completely novel task, the Digit Matrices. Notably, the Digit Matrices involve counterfactual reorderings of the most common numerical sequence (digits 1-9). GPT-3 outperformed human participants on this text-based matrix reasoning task, and displayed several behavioral signatures consistent with human-like reasoning. We also investigated several novel letter-string analogy variants, including an especially challenging set of problems involving zero-shot generalization from letter strings to real-world concepts (e.g., p q r : p q s :: cold cool warm : ? ). It is simply not plausible that these specific materials, or highly similar ones, were already present on the internet prior to our study. Accordingly, the ability of language models to solve these and many other problems (see Supplementary Section S3 of the original paper~\cite{webb2023emergent} for discussion of our other test materials), at a level on par with human participants, cannot be easily dismissed as simple mimicry of the training data.

\section{Counterfactual tasks}

\captionsetup{labelfont=bf,font=small}
\begin{figure}[h!]
\centering
\begin{subfigure}[t]{0.02\textwidth}
    \textbf{\fontfamily{phv}\selectfont{a}}
\end{subfigure}
\begin{subfigure}[t]{.45\linewidth}\vskip 0pt
    \includegraphics[width=\linewidth]{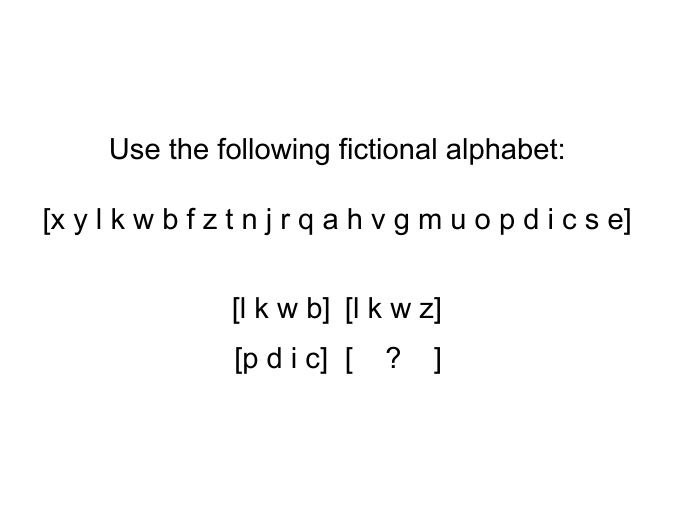} 
    \subcaption{}
    \label{task_example}
\end{subfigure}
\begin{subfigure}[t]{0.02\textwidth}
    \textbf{\fontfamily{phv}\selectfont{b}}
\end{subfigure}
\begin{subfigure}[t]{.45\linewidth}\vskip 0pt
    \includegraphics[width=\linewidth]{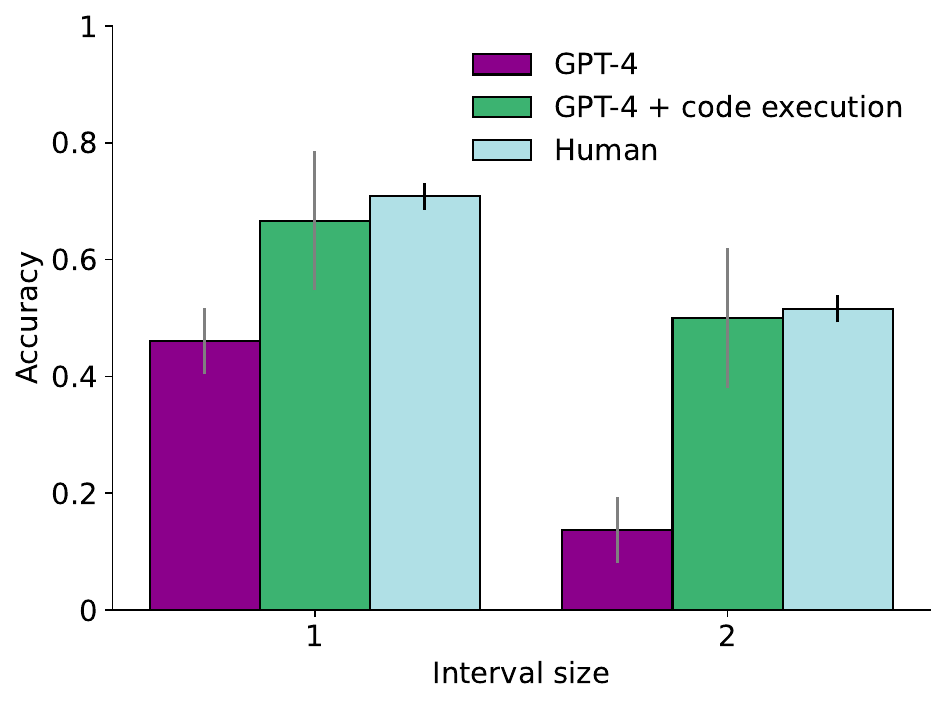} 
    \subcaption{}
    \label{results}
\end{subfigure}
\caption{\textbf{Results for letter-string analogies with shuffled alphabet.} \textbf{(a)} Example problem from Hodel \& West~\cite{hodel2023response}. Letter-string analogies are constructed based on a permuted alphabet. This example involves a successor relation with an interval size of 1, applied to the final letter of the string. Other problems involved an interval size of 2. See Methods (Supplementary Section S1) for more details. \textbf{(b)} Results for human participants, GPT-4, and a variant of GPT-4 augmented with the capacity to write and execute code (which the model used to identify the positions of letters in the permuted alphabet). Both humans and GPT-4 showed greater difficulty on problems involving an interval of size 2 (main effect of interval size, human participants: $P=1.5\times10^{-11}$, GPT-4: $P<2\times10^{-16}$). This effect did not reach significance for GPT-4 + code execution ($P=0.066$). Human participants outperformed GPT-4 (main effect of human participants vs. GPT-4: $P=8.9\times10^{-13}$), but GPT-4 + code execution performed on par with human participants (main effect of human participants vs. GPT-4 + code execution: $P=0.496$). When correct, GPT-4's responses were also accompanied by accurate explanations of the underlying rule, and incorrect responses were often based on a valid alternative rule (see Supplementary Results). Human results reflect average performance for $N=99$ participants for interval-size-1 and $N=97$ separate paticipants for interval-size-2. Black error bars represent standard error of the mean across participants. Grey error bars represent 95\% binomial CIs for average performance across multiple problems.} 
\label{results_fig}
\end{figure}

HW also present evidence from a specific variant of the letter-string analogy task, involving both a permuted alphabet and a larger interval size (Figure~\ref{task_example}), on which GPT-3 and GPT-4 display degraded performance. A similar task was also investigated by Lewis \& Mitchell (LM)~\cite{lewis2024using}, who similarly found that language models display worse performance on these problems than on those involving the standard alphabet. LM refer to this as a `counterfactual task' -- a term that was recently introduced to refer to unusual task variants, designed to deviate from common problems on which language models may have been trained ~\cite{wu2023reasoning}. HW and LM argue that degraded performance on this particular task variant suggests that the ability of language models to solve other types of analogy problems is based only on the similarity of those problems to the training data. This conclusion, however, ignores alternative potential explanations for the poor performance of these models on this particular task variant. Most notably, such problems require that letters be converted into the corresponding indices in the permuted alphabet, a process that depends on the ability to precisely count the items in a list. It is well known that language models have difficulty with counting, a phenomenon that may in fact be related to the capacity-limited nature of rapid numerical estimation in visual displays by humans (i.e., `subitizing')~\cite{kaufman1949discrimination,frankland2021no}. Thus, one potential alternative interpretation of these results is that they reflect not a general inability to perform analogical reasoning, but simply a specific difficulty with problems that require counting.

To test this alternative interpretation, we evaluated a more recently released variant of GPT-4, augmented with the capacity to write and execute code. This code execution capacity is often invoked by the model when performing tasks that require precise indexing within a list. Consistent with our alternative interpretation, GPT-4 was able to solve these `counterfactual' letter-string analogies at a roughly human level of performance when given the ability to count using code execution (Figure~\ref{results}), whereas without this functionality GPT-4 performed significantly worse, on par with the results of HW and LM.

Importantly, GPT-4 only relied on code execution to convert letters into their corresponding indices, using code that GPT-4 generated on its own. Moreover, it was not necessary to instruct GPT-4 to use code execution in this manner, nor to provide any task-specific instructions (e.g., regarding the importance of position or interval size). Furthermore, correct responses were typically accompanied by a coherent and accurate explanation (see example in Supplementary Results Section S2.2). In addition, many incorrect responses were based on a less abstract but nevertheless valid rule, at a rate similar to that observed for human participants (46\% of GPT-4's errors involved a valid alternative rule, compared with 39\% of errors in the human behavioral results reported by LM~\cite{lewis2024using}). Of course, the use of code execution to solve these particular problems is not a human-like solution (humans do not need to write code to count the items in a list). But these results suggest that the inability of language models to solve these particular reasoning problems stems from a specific difficulty with counting, rather than a general inability to solve analogy problems. Given the deliberate `counterfactual' design of this task (see Supplementary Results Section S2.4 for similar results from an additional counterfactual task), GPT-4's ability to solve these problems at a roughly human level, and to provide accurate explanations of its solutions, cannot be explained by the presence of these problems in the training data.

More generally, these results illustrate an important point about the evaluation of cognitive capacities in artificial systems. A central lesson of cognitive science is that cognition is comprised of interacting, but dissociable, processes. There is no particular reason to expect that these processes will similarly covary in artificial systems, especially those with radically different developmental origins from our own. Thus, as we previously argued~\cite{webb2023emergent}, it is important to disambiguate domain-specific failures in processes such as physical reasoning -- or, in this case, counting -- from the evaluation of core competencies such as analogical reasoning. Indeed, human reasoners also display dramatic variability in their capacity to deploy analogical reasoning in specific domains, based on their expertise in those domains~\cite{goldwater2021analogy}. This observation echoes recently articulated concerns about the potential confounding influence of auxiliary task demands when evaluating large language models~\cite{hu2024auxiliary}. Just as when testing young children or nonhuman animals, it is important to design evaluations that probe the capacity of interest while avoiding confounds resulting from auxiliary task demands.

\section{Mechanisms underlying emergent analogical reasoning}

A full account of emergent analogical reasoning in large language models, and of its relationship to human reasoning, will require us to go beyond behavioral evaluations by examining the internal mechanisms that support this capacity. While there is still much to learn, a growing body of evidence suggests that this capacity may be supported by a set of structured operations and emergent relational representations~\cite{todd2023function,hendel2023context,hernandez2023linearity,feng2023language}. Indeed, in-context learning -- the capacity of large language models to rapidly learn new tasks by conditioning their predictions on a small set of in-context task examples~\cite{brown2020language} -- may itself depend on analogical reasoning (or more generally, schema induction). This possibility is suggested by the dependence of in-context learning on the emergence of structured mechanisms for similarity-based inductive inference~\cite{olsson2022context,akyurek2024context}. Thus, the core mechanisms of few-shot learning and inference in transformer-based language models may depend on reasoning over patterns of similarity, mirroring the central role of similarity in human reasoning~\cite{gentner1989similarity,lu2022probabilistic,webb2023zero,webb2023relational}. It remains an important priority for future work to determine whether and how the mechanisms that support this capacity in large language models relate to those that implement reasoning in the human brain.

%TC:ignore

\section*{Acknowledgements}

This research has been supported by an award from Microsoft Azure as part of the Accelerating Foundation Models Research initiative.

\printbibliography

\pagebreak
\begin{refsection}
\begin{center}
\textbf{\LARGE Supplementary Materials}
\end{center}
\setcounter{figure}{0}
\setcounter{table}{0}
\setcounter{page}{1}
\setcounter{section}{0}
\makeatletter
\renewcommand{\figurename}{Supplementary Figure}
\renewcommand{\thesection}{\large S\arabic{section}} 
\renewcommand{\thepage}{S\arabic{page}}

\section{Methods}

\subsection{Code}

Code for all simulations can be downloaded from:

\begin{center}
    \href{https://github.com/taylorwwebb/counterfactual_analogies}{https://github.com/taylorwwebb/counterfactual\_analogies}
\end{center}

\noindent Most code was written in Python v3.11.4, using the following packages: NumPy v1.25.0~\cite{harris2020array}, SciPy v1.11.1~\cite{2020SciPy-NMeth}, statsmodels v0.14.0~\cite{seabold2010statsmodels}, and Matplotlib v3.7.1~\cite{Hunter:2007}. Logistic regression analyses were carried out in R v3.6.3~\cite{rcite}. Experimental stimuli for human behavioral experiments were written in JavaScript using jsPsych v7.2.1~\cite{de2015jspsych}.

\subsection{Problem Set}

Letter string analogy problems were created using the following permuted alphabet, also used by Hodel \& West (HW)~\cite{hodel2023response}:

\begin{center}
[x y l k w b f z t n j r q a h v g m u o p d i c s e]
\end{center}

\noindent Each letter string analogy problem involved one of six transformation types: sequence extension, successor, predecessor, removing a redundant letter, fixing a sequence, and sorting. These are the same transformation types used in our original study~\cite{webb2023emergent}. In the sequence extension transformation, the source involved an ordered sequence of four letters (based on the ordering in the permuted alphabet) followed by an extension of this sequence involving five letters (including the next letter in the permuted alphabet), as in the following example:

\begin{center}
[x y l k] [x y l k w]
\end{center}

\noindent In the successor transformation, the source involved an ordered sequence of four letters, followed by that same sequence, but with the final letter replaced by its successor in the permuted alphabet, as in the following example:

\begin{center}
[x y l k] [x y l w]
\end{center}

\noindent In the predecessor transformation, the source involved an ordered sequence of four letters, followed by that same sequence, but with the first letter replaced by its predecessor in the permuted alphabet, as in the following example:

\begin{center}
[y l k w] [x l k w]
\end{center}

\noindent In the transformation involving removal of a redundant letter, the source involved an ordered sequence of five letters with one letter repeated, followed by that same sequence with the redundant letter removed, as in the following example:

\begin{center}
[x y y l k w] [x y l k w]
\end{center}

\noindent In the transformation involving fixing a sequence, the source involved an ordered sequence of five letters with one out-of-place letter (violating the ordering in the permuted alphabet), followed by that same sequence with the out-of-place letter replaced, as in the following example:

\begin{center}
[x y l g w] [x y l k w]
\end{center}

\noindent In the sorting transformation, the source involved an ordered sequence of five letters with the position of two letters swapped, followed by a sorted version of the same sequence (sorted based on the ordering in the permuted alphabet), as in the following example:

\begin{center}
[x k l y w] [x y l k w]
\end{center}

\noindent Similar to HW, we only investigated problems referred to as `zero-generalization' in our original study, meaning that the only difference between source and target sequences was the use of different letters. An example problem (involving a successor relation) is presented below:

\begin{center}
[x y l k] [x y l w]
\linebreak[4][j r q a] [j r q h]
\end{center}

\noindent We investigated problems involving an interval size of either 1 or 2. The problems with an interval size of 1 are exactly as described above. The interval size of 2 was employed in different ways, depending on the transformation type. In the sequence extension transformation, the first sequence was presented as usual, but the extension of that sequence involved the addition of a letter separated by a successor interval of 2 in the permuted alphabet, as in the following example:

\begin{center}
[x y l k] [x y l k b]
\end{center}

\noindent Similarly, in the successor transformation, the final letter in the first sequence was replaced by a letter separated by a successor interval of 2, as in the following example:

\begin{center}
[x y l k] [x y l b]
\end{center}

\noindent In the predecessor transformation, the first letter in the sequence was replaced by a letter that preceded it by an interval of 2, as in the following example:

\begin{center}
[l k w b] [x k w b]
\end{center}

\noindent In the transformation involving removal of a redundant letter, fixing a sequence, and sorting, the source involved an ordered sequence of five letters, in which each letter was separated by an interval of 2 in the permuted alphabet. The following example illustrates this for the transformation involving removal of a redundant letter:

\begin{center}
[x l l w f t] [x l w f t]
\end{center}

\noindent The following example illustrates this for the transformation involving fixing a sequence:

\begin{center}
[x l w g t] [x l w f t]
\end{center}

\noindent The following example illustrates this for the sorting transformation:

\begin{center}
[x f w l t] [x l w f t]
\end{center}

\noindent Problems with an interval of size 2 were the same as those used by HW, and problems with an interval size of 1 were generated using code from HW, both downloaded from the following repository:

\begin{center}
    \href{https://github.com/hodeld/emergent_analogies_LLM_fork}{https://github.com/hodeld/emergent\_analogies\_LLM\_fork}
\end{center}

\noindent There were 600 problems with an interval size of 1, and 600 problems with an interval size of 2 (each involving 100 problems with each transformation type). The same problem sets were used for testing GPT-4 and human participants.

\subsection{Evaluating GPT-4}

We queried GPT-4 in an automated fashion using the OpenAI Chat Completions API. Experiments were performed using the `gpt-4-0125-preview' engine, with temperature and top-p set both set to 0. Prompts and analogy problems were presented using the `User' role, with the default message from the `System' role (`You are a helpful assistant.'). The prompt format that we used is illustrated in the following example:

\begin{center}
    Use this fictional alphabet: [x y l k w b f z t n j r q a h v g m u o p d i c s e].
    \linebreak[4]
    \linebreak[4]Let's try to complete the pattern:
    \linebreak[4]
    \linebreak[4][x y l k] [x y l w]
    \linebreak[4][j r q a] [ ? ]
    \linebreak[4]
    \linebreak[4]Please only provide the answer. Do not provide any additional explanation.
    \linebreak[4]
    \linebreak[4]Answer:
\end{center}

\noindent This prompt was employed to ensure that responses adhered to a regular format, thus avoiding the difficulty involved in parsing the verbose responses that GPT-4 typically provides. Line breaks were indicated using the `\textbackslash n' character. We evaluated GPT-4 on 300 problems with an interval size of 1, and 300 problems with an interval size of 2 (each involving 50 problems with each transformation type).

\subsection{Evaluating GPT-4 with code execution}

We also evaluated a variant of GPT-4 that was augmented with a capacity to write and execute code. These experiments were performed with the OpenAI Assistants API, using the same engine (`gpt-4-0125-preview') as the experiment without code execution. Temperature and top-p were both set to their default values of 1. The prompt format that we used is illustrated in the following example:

\begin{center}
    Let's solve a puzzle problem involving the following fictional alphabet:
    \linebreak[4]
    \linebreak[4][x y l k w b f z t n j r q a h v g m u o p d i c s e]
    \linebreak[4]
    \linebreak[4]Here is the problem:
    \linebreak[4]
    \linebreak[4][x y l k] [x y l w]
    \linebreak[4][j r q a] [ ? ]
\end{center}

\noindent This is the only prompt that was provided to GPT-4. We did not provide any specific instructions to use code execution to solve the problem, nor did we indicate the importance of the position or interval size between letters in the permuted alphabet. GPT-4 typically responded with an extended `chain-of-thought' response~\cite{wei2022chain}, often invoking the code execution function. Code execution was typically invoked to identify the corresponding position of the letters in the permuted alphabet. 

GPT-4 sometimes refused to provide an answer, insisting that there was too much uncertainty about the pattern underlying the analogy problem. In these cases, GPT-4's response was re-generated until it provided an answer. When GPT-4 did provide an answer, this was counted as final, even if GPT-4 expressed uncertainty. 

GPT-4's chain-of-thought responses typically had a highly variable format, making it difficult to automate evaluation. We therefore manually inspected the responses to determine accuracy. Due to the time involved in manually assessing these responses, these experiments involved a smaller subset of 60 problems with an interval size of 1, and 60 problems with an interval size of 2 (each involving 10 problems with each transformation type). GPT-4's complete response, including any code generated, was recorded.

\subsection{Human Behavioral Experiment}

Human behavioral data was collected in an online experiment. The experiment was approved by the UCLA Institutional Review Board (IRB protocol \#22-000841, approved May 17, 2022), and all participants provided informed consent. Participants were recruited through the Prolific platform. Ninety-nine participants (32 female, 19-60 years old, average age = 29 years old) completed the experiment in the interval-size-1 condition, and ninety-seven participants (40 female, 18-54 years old, average age = 29 years old) completed the experiment in the interval-size-2 condition. All participants spoke fluent English and had normal or corrected-to-normal vision. No statistical methods were used to pre-determine sample sizes. Participants were paid \$3.00 to participate, and the experiment took a maximum of 10 minutes to complete.

Participants were first presented with a set of instructions, and the following example problem (not involving the permuted alphabet or any of the transformations employed in the actual experiment):

\begin{center}
[a a a] [b b b]
\linebreak[4][c c c] [~ ? ~]
\end{center}

Each participant completed 6 problems, including 1 problem for each transformation type. The specific problem instances were randomly sampled for each participant, and participants received these problems in a random order. Participants generated a free response for each problem. Participants performed the experiment in full-screen mode, and were instructed to turn off any music or other potential sources of distraction. After completing the problems, participants completed an attention check involving a simple categorical judgment (they were presented with a list of items and asked to identify which one was categorized as a vegetable). All participants successfully completed the attention check.

\subsubsection{Statistical analyses}

Logistic regression analyses were carried out at the individual trial level, with each data point corresponding to a particular trial from a particular participant (or GPT-4). The dependent variable in all regression analyses was a binary variable coding for whether a particular response was correct or incorrect. 

One regression analysis compared human participants with GPT-4. This analysis included three predictors: a binary predictor coding for interval size (1 vs. 2), a binary predictor coding for GPT-4 vs. human participants, and an interaction term. A second regression analysis compared human participants with GPT-4 + code execution, using the same predictors. Three separate regression analyses were then performed looking only at data from human participants, GPT-4, or GPT-4 + code execution, with a single binary predictor coding for interval size.

\section{Supplementary Results}

\subsection{Data Availability}

Data for the human behavioral experiment and GPT-4 evaluations, along with the letter string problem sets, can be downloaded from:

\begin{center}
    \href{https://github.com/taylorwwebb/counterfactual_analogies}{https://github.com/taylorwwebb/counterfactual\_analogies}
\end{center}

\subsection{Example GPT-4 response involving code execution}

The following example illustrates the behavior of GPT-4 when augmented with the ability to execute code. This example is in response to the following analogy problem:

\begin{center}
    Let's solve a puzzle problem involving the following fictional alphabet:
    \linebreak[4]
    \linebreak[4][x y l k w b f z t n j r q a h v g m u o p d i c s e]
    \linebreak[4]
    \linebreak[4]Here is the problem:
    \linebreak[4]
    \linebreak[4][q a h v] [q a h m]
    \linebreak[4][k w b f] [ ? ]
\end{center}

\noindent This problem involves a successor relation with an interval size of 2. The correct answer is [k w b t]. GPT-4's response is as follows (typewriter font indicates code generated and executed by GPT-4):

\begin{quote}
    This puzzle appears to revolve around matching sequences of characters from the provided fictional alphabet. To solve the puzzle, we'll need to determine the relationship between the given sequences and then apply that relationship to find the missing sequence. Let's identify the pattern or rule that connects the first pair of sequences ([q a h v] and [q a h m]) and then use this pattern to solve for the missing sequence corresponding to [k w b f].

    Let's go through the fictional alphabet and the given sequences to identify the pattern.

    Looking at the sequences:

    1. ([q a h v]) changes to ([q a h m]).

    By examining the given fictional alphabet, let's find the positioning of each character and identify the transformation applied from the first sequence to the second:

    Fictional Alphabet: ([x y l k w b f z t n j r q a h v g m u o p d i c s e])

    Let's start by determining the index of each character in the first sequence and then see how it transforms into the second sequence.

    \texttt{\# Fictional alphabet as a list of characters for indexing purposes}
    \linebreak[4]\texttt{fictional\_alphabet = [`x', `y', `l', `k', `w', `b', `f', `z', `t', `n', `j', `r', `q', `a', `h', `v', `g', `m', `u', `o', `p', `d', `i', `c', `s', `e']}

    \texttt{\# Given sequences}
    \linebreak[4]\texttt{sequence1 = [`q', `a', `h', `v']}
    \linebreak[4]\texttt{sequence2 = [`q', `a', `h', `m']}

    \texttt{\# Let's find the indexes of each character in both sequences}
    \linebreak[4]\texttt{sequence1\_indexes = [fictional\_alphabet.index(char) for char in sequence1]}
   \linebreak[4]\texttt{sequence2\_indexes = [fictional\_alphabet.index(char) for char in sequence2]}

   \texttt{sequence1\_indexes, sequence2\_indexes}

    The indexes of the characters in the first sequence ([q a h v]) are ([12, 13, 14, 15]) and in the second sequence ([q a h m]) are ([12, 13, 14, 17]).

    Observing the transformation from the first sequence to the second, it looks like all but the last item remain the same. The last character changes from having an index of (15) to an index of (17).

    Now let's apply this observed transformation rule to the sequence ([k w b f]) and find the corresponding missing sequence. We will maintain the first three characters and find the character two positions after the last character in the fictional alphabet.

    \texttt{\# Sequence to transform}
    \linebreak[4]\texttt{sequence\_to\_transform = [`k', `w', `b', `f']}

    \texttt{\# Getting indexes of the characters in the sequence to transform}
    \linebreak[4]\texttt{transform\_indexes = [fictional\_alphabet.index(char) for char in sequence\_to\_transform]}

    \texttt{\# Change the last character as per observed transformation: 2 positions after the original}
    \linebreak[4]\texttt{new\_last\_char\_index = (transform\_indexes[-1] + 2) \% len(fictional\_alphabet)  \# Using modulo in case of overflow}

    \texttt{\# create the new sequence with the transformation applied}
     \linebreak[4]\texttt{new\_sequence = sequence\_to\_transform[:3] + [fictional\_alphabet[new\_last\_char\_index]]}

    \texttt{new\_sequence}

    The missing sequence corresponding to ([k w b f]) after applying the observed transformation rule is ([k w b t]). 

    So the solution to the puzzle is:
    \begin{itemize}
        \item{([q a h v]) transforms to ([q a h m])}
        \item{([k w b f]) transforms to ([k w b t])}
    \end{itemize}
        
\end{quote}

\noindent A number of observations can be made about this response. First, GPT-4 spontaneously uses chain-of-thought reasoning~\cite{wei2022chain} to decompose the problem, developing a plan that involves first identifying the change between the first and second sequences, and then applying that change to the third sequence. GPT-4 then uses code execution to convert the letters in the source sequences to their indices in the fictional alphabet, and identifies the rule based on these indices (that the first three letters stay the same, while the last letter is incremented by an interval of 2). Finally, GPT-4 again uses code execution to apply this rule to the target sequence and generate an answer. GPT-4's responses typically involved a similar combination of chain-of-thought reasoning and code execution, and correct answers were always accompanied by an accurate explanation of the underlying rule. Incorrect answers were often accompanied by an analysis that identified a valid alternative rule. The complete set of responses is available at:

\begin{center}
    \href{https://github.com/taylorwwebb/counterfactual_analogies/gpt-4-0125-preview_code_execution}{https://github.com/taylorwwebb/counterfactual\_analogies/gpt-4-0125-preview\_code\_execution}
\end{center}

\pagebreak

\subsection{Results by problem type}

\captionsetup{labelfont=bf,font=small}
\begin{figure}[h!]
\centering
\begin{subfigure}[t]{0.02\textwidth}
    \textbf{\fontfamily{phv}\selectfont{a}}
\end{subfigure}
\begin{subfigure}[t]{.8\linewidth}\vskip 0pt
    \includegraphics[width=\linewidth]{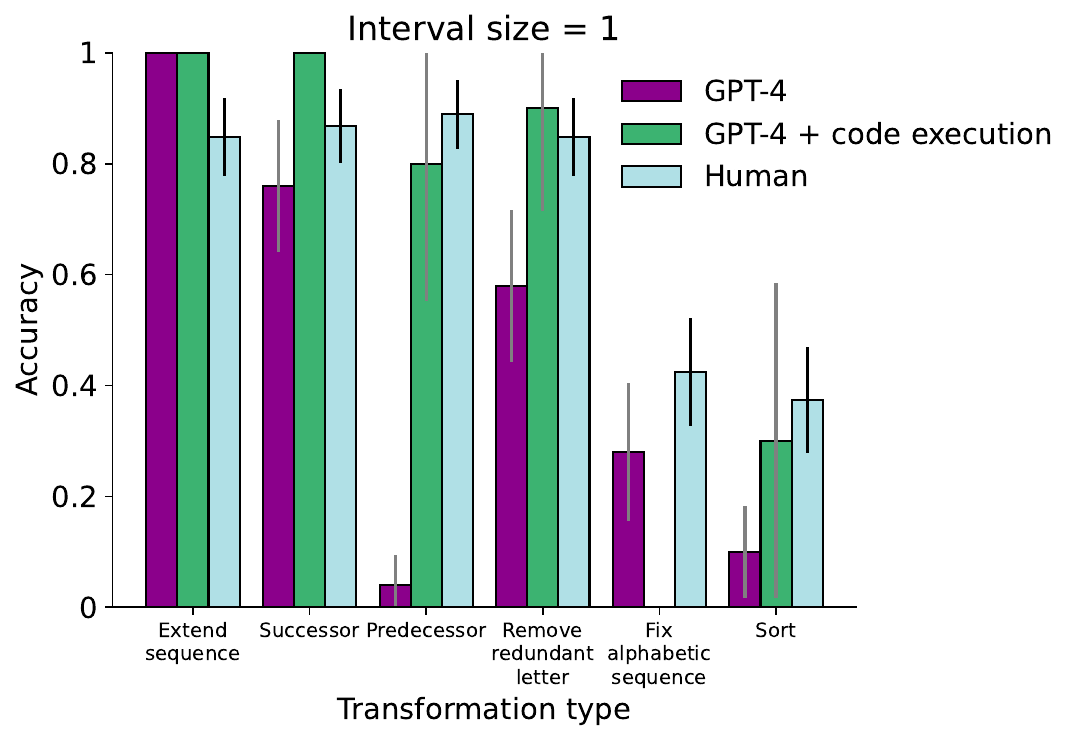} 
    \subcaption{}
    \label{int1_results_fig}
\end{subfigure}
\\
\begin{subfigure}[t]{0.02\textwidth}
    \textbf{\fontfamily{phv}\selectfont{b}}
\end{subfigure}
\begin{subfigure}[t]{.8\linewidth}\vskip 0pt
    \includegraphics[width=\linewidth]{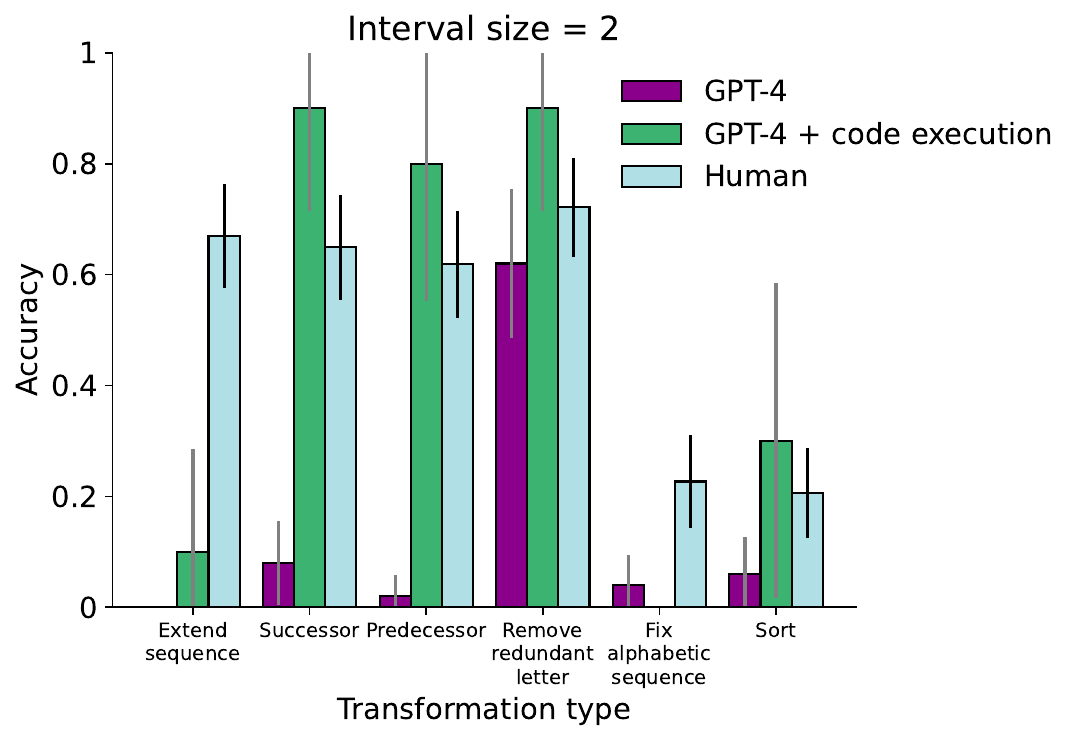} 
    \subcaption{}
    \label{int2_results_fig}
\end{subfigure}
\caption{\textbf{Results for letter-string analogies with shuffled alphabet, sorted by transformation type and interval size.} \textbf{(a)} Results for human participants ($N=99$), GPT-4, and GPT-4 + code execution on problems with an interval size of 1. \textbf{(b)} Results for human participants ($N=97$), GPT-4, and GPT-4 + code execution on problems with an interval size of 2. Black error bars represent standard error of the mean across participants. Grey error bars represent 95\% binomial CIs for average performance across multiple problems.} 
\label{detailed_results_fig}
\end{figure}

\pagebreak

Supplementary Figure~\ref{detailed_results_fig} show results broken down by problem type. GPT-4 + code execution matched the pattern of human performance reasonably well, although the sample size for each problem type is small. A notable exception was for problems involving the extend-sequence transformation and an interval size of 2, where GPT-4 + code execution answered 9/10 problems incorrectly. On these problems, GPT-4 did not invoke code execution to identify the interval between the letters in the source sequences, and mistakenly identified an interval size of 1, then using the wrong interval size to complete the analogy. This performance contrasts with the one problem of this type that GPT-4 answered correctly, for which GPT-4 correctly identified the source interval using code execution. These findings further underscore the extent to which GPT-4's performance on these problems is constrained by its difficulty with counting.

\subsection{Evaluating the impact of training data cutoff date}

The GPT-4 engine employed in these experiments (`gpt-4-0125-preview') has a training data cutoff date of December 2023. This is after the publication of HW's commentary~\cite{hodel2023response}, which included the problem set used in these experiments, raising the concern that these problems might in principle be present in the training data for the version of GPT-4 used in our experiments. It is worth noting first that this possibility could not explain how GPT-4 was able to provide accurate explanations for its responses, nor how it was able to generate code for solving these problems, neither of which are present in the publicly available dataset. However, we also performed two additional experiments to directly address this concern. First, we evaluated GPT-4 + code execution with a novel problem set using the following alternative synthetic alphabet:

\begin{center}
[n h v b o p y z t m r w x f i q d j l c a s k g e u]
\end{center}

\begin{figure}[h!]
\centering
\begin{subfigure}[t]{0.02\textwidth}
    \textbf{\fontfamily{phv}\selectfont{a}}
\end{subfigure}
\begin{subfigure}[t]{.75\linewidth}\vskip 0pt
    \includegraphics[width=\linewidth]{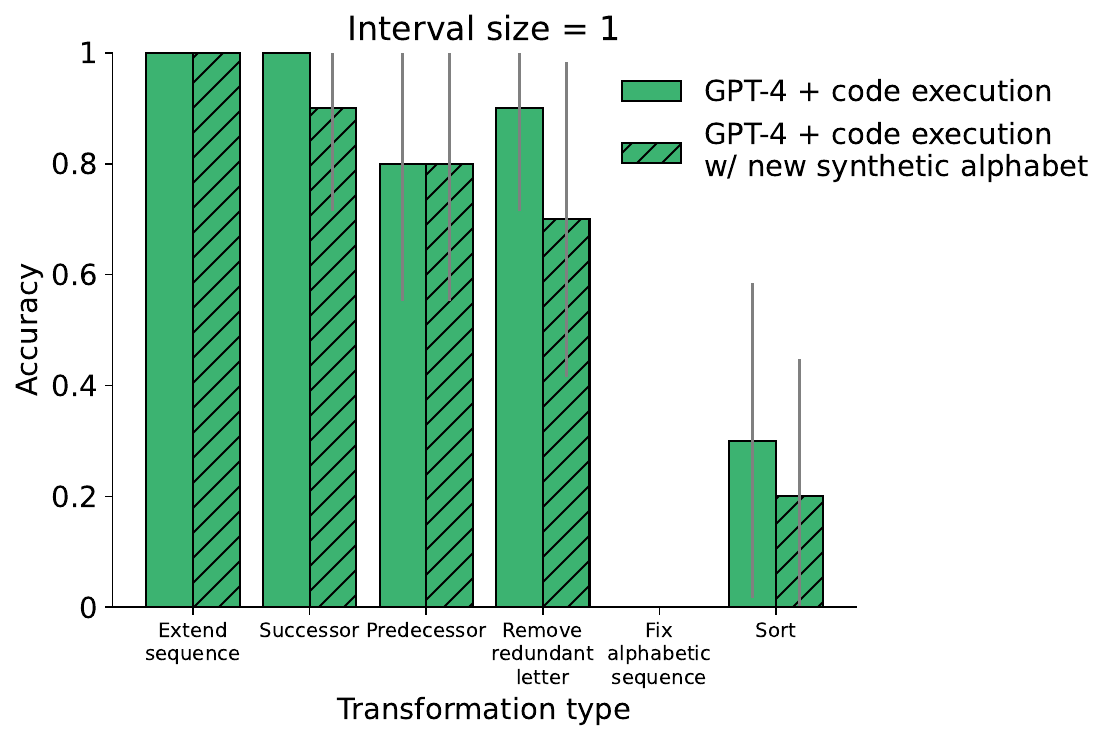} 
    \subcaption{}
    \label{int1_results_comparing_alphabets_fig}
\end{subfigure}
\\
\begin{subfigure}[t]{0.02\textwidth}
    \textbf{\fontfamily{phv}\selectfont{b}}
\end{subfigure}
\begin{subfigure}[t]{.75\linewidth}\vskip 0pt
    \includegraphics[width=\linewidth]{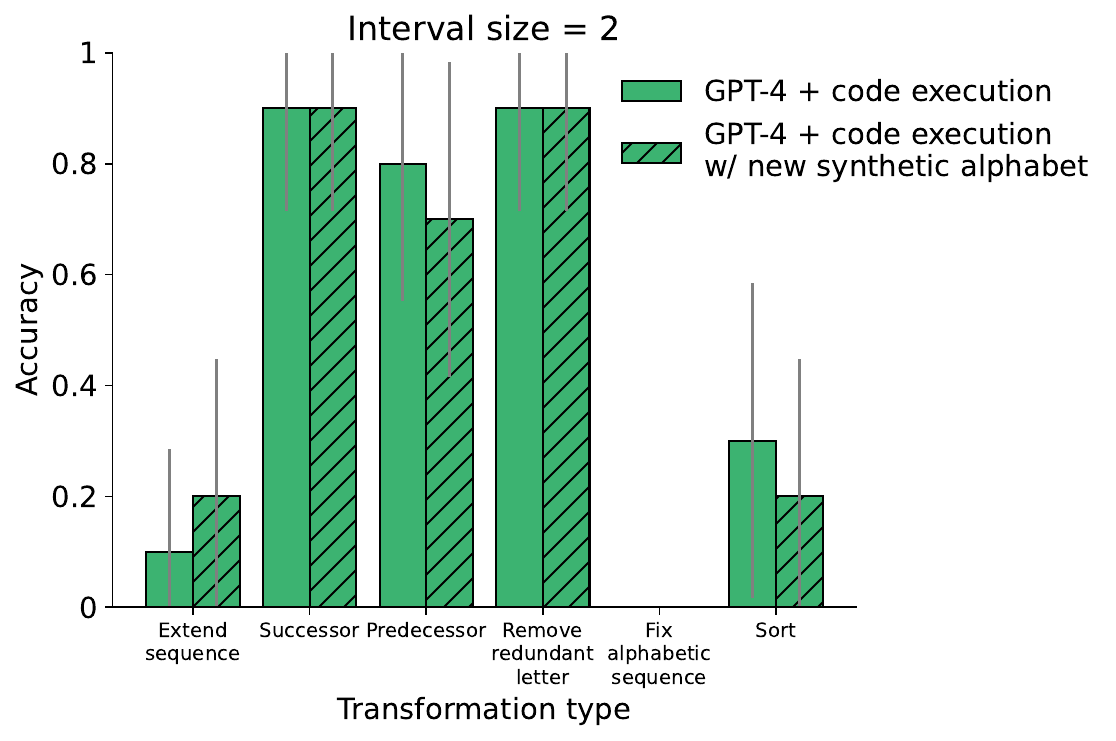} 
    \subcaption{}
    \label{int2_results_comparing_alphabets_fig}
\end{subfigure}
\caption{\textbf{Results for GPT-4 + code execution on the original vs. new synthetic alphabets.} \textbf{(a)} Results for problems with an interval size of 1. \textbf{(b)} Results for problems with an interval size of 2. Performance was comparable for the two alphabets (logistic regression with binary predictor coding for old vs. new alphabet: $P=0.52$). Error bars represent 95\% binomial CIs for average performance across multiple problems.} 
\label{old_vs_new_alphabet_results}
\end{figure}

\begin{figure}[h!]
\centering
\begin{subfigure}[t]{0.02\textwidth}
    \textbf{\fontfamily{phv}\selectfont{a}}
\end{subfigure}
\begin{subfigure}[t]{.6\linewidth}\vskip 0pt
    \includegraphics[width=\linewidth]{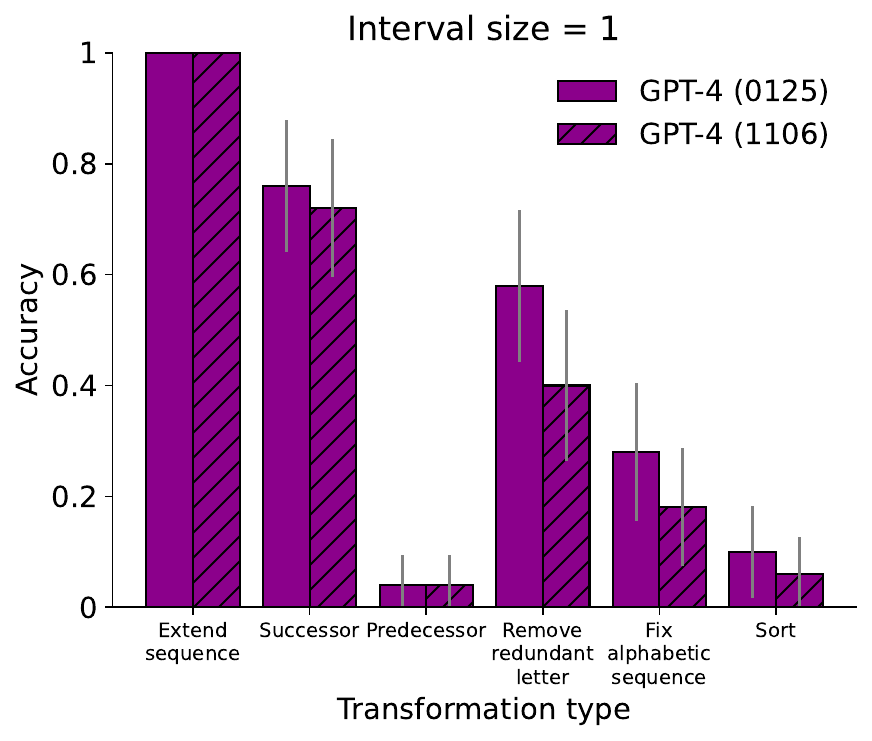} 
    \subcaption{}
    \label{int1_results_comparing_engines_fig}
\end{subfigure}
\\
\begin{subfigure}[t]{0.02\textwidth}
    \textbf{\fontfamily{phv}\selectfont{b}}
\end{subfigure}
\begin{subfigure}[t]{.6\linewidth}\vskip 0pt
    \includegraphics[width=\linewidth]{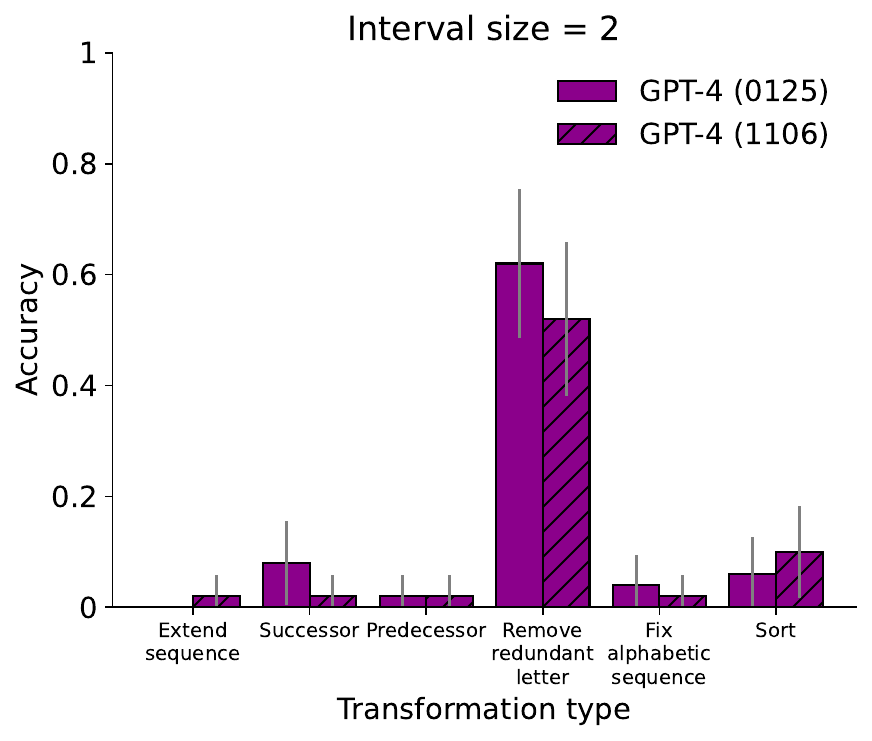} 
    \subcaption{}
    \label{int2_results_comparing_engines_fig}
\end{subfigure}
\caption{\textbf{Results for GPT-4 using engines with different training data cutoff dates.} \textbf{(a)} Results for problems with an interval size of 1. \textbf{(b)} Results for problems with an interval size of 2. Performance was comparable for the two engines (logistic regression with binary predictor coding for old vs. new engine: $P=0.12$). Error bars represent 95\% binomial CIs for average performance across multiple problems.} 
\label{old_vs_new_engine_results}
\end{figure}

\noindent Performance on these problems was nearly identical to the performance observed for the original synthetic alphabet (Supplementary Figure~\ref{old_vs_new_alphabet_results}). Second, we performed the experiments on GPT-4 without code execution using an older GPT-4 engine, `gpt-4-1106-preview', with a training data cutoff date of April 2023, prior to the publication of HW's commentary and problem set. Performance was very similar for the two engines (Supplementary Figure~\ref{old_vs_new_engine_results}). Thus, performance was driven almost entirely by the presence or absence of code execution. With code execution, GPT-4 performed well on both the original problems and new problems with a unique alphabet; whereas without code execution, GPT-4 performed poorly, regardless of the training data cutoff date. This pattern of results is inconsistent with the presence of these problems in GPT-4's training data.

\pagebreak

\captionsetup{labelfont=bf,font=small}

\pagebreak

\captionsetup{labelfont=bf,font=small}

\pagebreak

\subsection{Errors based on alternate rules}

For some problem types, it is possible to define an alternative rule that validly describes the problems. These alternative rules are typically less abstract than the intended rule. For example, consider the following example problem, based on a sort rule:

\begin{center}
    Fictional alphabet: [x y l k w b f z t n j r q a h v g m u o p d i c s e].
    \linebreak[4]
    \linebreak[4][x k l y w] [x y l k w]
    \linebreak[4][h r q a j] [ ? ]
\end{center}

\noindent In this problem, the second sequence ([x y l k w]) can be obtained by sorting the first sequence ([x k l y w]) according to the ordering in the permuted alphabet. The correct answer is [j r q a h], which is obtained by applying this rule to the third sequence ([h r q a j]). However, an alternative interpretation is that the rule involves swapping the second and fourth entries in a sequence. Applying this alternative rule yields the answer [h a q r j].

We quantified the number of errors made by GPT-4 + code execution that could be classified as involving a valid alternative rule (Tables~\ref{int1_valid_errors} and~\ref{int2_valid_errors}). Errors based on alternative rule use were especially common for problems involving fixing an alphabetic sequence or sorting. Altogether, errors based on alternative rules constituted 46\% (23/50) of GPT-4's errors, consistent with the rate observed in the human behavioral data from Lewis \& Mitchell~\cite{lewis2024using}. This result suggests that, in addition to improving overall performance, the ability to count using code execution also leads GPT-4 to make more human-like errors.

\begin{table}[!h]
\begin{center}
\caption{\textbf{Fraction of errors involving valid alternative rules, interval size = 1}} 
\label{int1_valid_errors}
\begin{tabular}{|c|c|} 
 \hline
 \textbf{Transformation type} & \textbf{Valid errors} \\ [0.5ex] 
 \hline\hline
 Extend sequence & 0\textbackslash 0 \\ 
 \hline
 Successor & 0\textbackslash 0 \\ 
 \hline
 Predecessor & 0\textbackslash 2 \\ 
 \hline
 Remove redundant letter & 1\textbackslash 1 \\ 
 \hline
 Fix alphabetic sequence & 6\textbackslash 10 \\ 
 \hline
 Sort & 3\textbackslash 7 \\ 
 \hline
\end{tabular}
\end{center}
\end{table}

\begin{table}[!h]
\begin{center}
\caption{\textbf{Fraction of errors involving valid alternative rules, interval size = 2}} 
\label{int2_valid_errors}
\begin{tabular}{|c|c|} 
 \hline
 \textbf{Transformation type} & \textbf{Valid errors} \\ [0.5ex] 
 \hline\hline
 Extend sequence & 0\textbackslash 9 \\ 
 \hline
 Successor & 0\textbackslash 1 \\ 
 \hline
 Predecessor & 0\textbackslash 2 \\ 
 \hline
 Remove redundant letter & 0\textbackslash 1 \\ 
 \hline
 Fix alphabetic sequence & 10\textbackslash 10 \\ 
 \hline
 Sort & 3\textbackslash 7 \\ 
 \hline
\end{tabular}
\end{center}
\end{table}

\pagebreak

\printbibliography[title={Supplementary References}]

\end{refsection}

%TC:endignore

\end{document}